\documentclass[fleqn,10pt,twocolumn]{wlscirep}
\usepackage[utf8]{inputenc}
\usepackage[T1]{fontenc}
\usepackage{amsmath}
\usepackage{amssymb}
\usepackage{csquotes}
\usepackage{booktabs}
\usepackage{dblfloatfix}

\usepackage{subcaption}

\title{Quantification of Robotic Surgeries with Vision-Based Deep Learning}

\author[1,*]{Dani Kiyasseh}
\author[3]{Runzhuo Ma}
\author[3]{Taseen F. Haque}
\author[3]{Jessica Nguyen}
\author[2]{Christian Wagner}
\author[1]{Animashree Anandkumar}
\author[3]{Andrew J. Hung}
\affil[1]{Department of Computing and Mathematical Sciences, California Institute of Technology, California, USA}
\affil[2]{Department of Urology, Pediatric Urology and Uro-Oncology, Prostate Center Northwest, St. Antonius-Hospital, Gronau, Germany}
\affil[3]{Center for Robotic Simulation and Education, Catherine \& Joseph Aresty Department of Urology, University of Southern California, California, USA}

\affil[*]{dkiyass1@caltech.edu}

\begin{abstract}
Surgery is a high-stakes domain where surgeons must navigate critical anatomical structures and actively avoid potential complications while achieving the main task at hand. Such surgical activity has been shown to affect long-term patient outcomes. To better understand this relationship, whose mechanics remain unknown for the majority of surgical procedures, we hypothesize that the core elements of surgery must first be quantified in a reliable, objective, and scalable manner. We believe this is a prerequisite for the provision of surgical feedback and modulation of surgeon performance in pursuit of improved patient outcomes. To holistically quantify surgeries, we propose a unified deep learning framework, entitled Roboformer, which operates exclusively on videos recorded during surgery to independently achieve multiple tasks: surgical phase recognition (the \textit{what} of surgery), gesture classification and skills assessment (the \textit{how} of surgery). We validated our framework on four video-based datasets of two commonly-encountered types of steps (dissection and suturing) within minimally-invasive robotic surgeries. We demonstrated that our framework can generalize well to unseen videos, surgeons, medical centres, and surgical procedures. We also found that our framework, which naturally lends itself to explainable findings, identified relevant information when achieving a particular task. These findings are likely to instill surgeons with more confidence in our framework's behaviour, increasing the likelihood of clinical adoption, and thus paving the way for more targeted surgical feedback. 
\end{abstract}

\begin{document}

\flushbottom
\maketitle
%
%
\thispagestyle{empty}

\section*{Introduction}

Surgery is an inherently high-stakes domain in which surgeons must navigate critical anatomical structures such as vasculature and nerves, avoid harming healthy tissue, and actively avoid potential complications, all the while tending to the main task at hand. In contrast to this complexity, the overarching goal of surgery is quite straightforward; to improve post-operative surgical and patient outcomes. Although the determinants of post-operative patient outcomes are multi-factorial, recent studies have demonstrated the relationship between intra-operative surgical activity (\textit{what} and \textit{how} a surgical procedure is performed) and long-term patient outcomes\cite{Birkmeyer2013}. A better understanding of this relationship, whose mechanics remain unknown for the majority of surgical procedures, would shed light, for example, on what and how surgical behaviour can be modulated to ultimately improve patient outcomes.

More broadly, to better understand the relationship between intra-operative surgical activity and post-operative patient outcomes, we hypothesize that the core elements of surgical procedures must first be quantified in a reliable, objective, and scalable manner. Such a holistic quantification of surgery, which has yet to be achieved, is a prerequisite to guide the provision of surgical feedback, modulate surgeon performance, and positively influence patient outcomes. To that end, we break down the core elements of surgery into; (a) \textit{what} step, or phase, of the surgical procedure is being performed and (b) \textit{how} that step is executed by the surgeon, as outlined next. 

\begin{figure*}[!t]
    \centering
    \begin{subfigure}{\textwidth}
    \centering
    \includegraphics[width=0.8\textwidth]{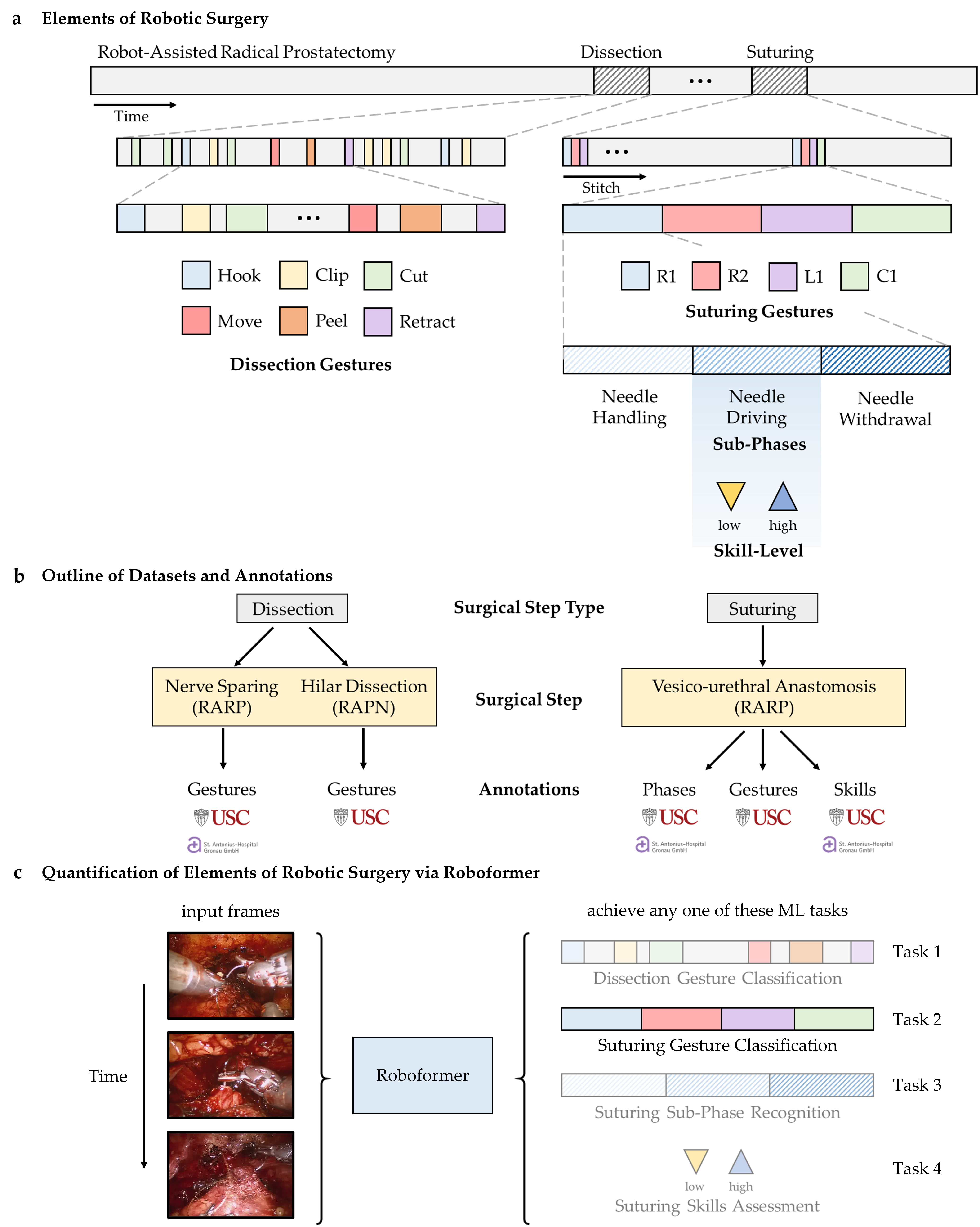}
    \end{subfigure}
    \caption{\small \textbf{Elements and quantification of a robotic surgery.} \textbf{(a)} Many surgeries, including a robot-assisted radical prostatectomy (RARP), entail dissection and suturing steps. A particular dissection step is composed of a sequence of distinct actions, or gestures, that are performed by surgeons. A suturing step is composed of multiple stitches, each of which is performed via a particular suturing gesture. Each stitch can also be broken down into three sub-phases (needle handling, driving, and withdrawal) and assessed for their quality (low vs. high). \textbf{(b)} Our data span videos, from two medical institutions, of two surgical step types and three surgery-specific steps, and corresponding human-generated annotations. \textbf{(c)} To quantify the elements of surgery, our vision-based deep learning framework (Roboformer) receives an input of live surgical frames and achieves any one of Task 1-4. These range from surgical phase recognition to gesture classification and skills assessment. We highlight a particular task (Task 2) to emphasize that we achieve a single task at any one point and not all simultaneously.}
    \label{fig:datasets_and_tasks}
\end{figure*}

Despite efforts to consolidate the definition of different elements of a surgical procedure\cite{Meireles2021}, a surgical step is typically thought of as a segment necessary for the completion of a surgical procedure. For example, a robot-assisted radical prostatectomy (RARP), a procedure in which the prostate gland is removed from a patient's body, often consists of twelve main steps\cite{Skarecky2013,Lovegrove2016}. We depict two types of steps (tissue dissection and suturing) in Fig.~\ref{fig:datasets_and_tasks}a. By quantifying the \textit{what} of surgery, surgeons can better view the duration and ordering of surgical steps within a particular procedure, whether that deviated from the typical approach, and, if so, allow for post-hoc video inspection of such steps. 

While high-level surgical steps, and their sequence over time, are often known and fixed prior to surgery, \textit{how} these steps are executed by surgeons can differ and be viewed through two lenses; that of actions and technical skill. For example, to complete a dissection step within the RARP procedure, surgeons perform a sequence of discrete actions, or gestures, via robotic arms (see Fig.~\ref{fig:datasets_and_tasks}a left column). It has been shown that novice and experienced surgeons employ distinct sequences of gestures\cite{Reiley2009}, implying perhaps that an optimal gesture sequence may exist, and thus be learned, for a particular surgical step. In addition to manifesting as a sequence of gestures, a surgical step can be executed with a varying degree of technical skill\cite{Hamdorf2000}. For example, the suturing step within the RARP procedure can be identified as reflecting low- or high-skill (see Fig.~\ref{fig:datasets_and_tasks}a right column). There is evidence to suggest that high surgeon skill levels are associated with lower rates of serious morbidity\cite{Stulberg2020,Curtis2020} and higher rates of 5-year patient survival\cite{Brajcich2021}. Therefore, the quantification of skills assessment opens the door for targeted feedback during surgical training as a way to influence long-term patient outcomes.


Previous research has often quantified surgical activity through the use of deep learning, a subfield of artificial intelligence in which patterns are learned from data\cite{Hashimoto2018,Maier2017}. One line of research has focused on exploiting robot-derived sensor data, such as the displacement and velocity of the robotic arms (kinematics), to predict clinical outcomes\cite{Weede2012,Hung2018APM1,Hung2018APM2,Hung2019}. For example, Hung \textit{et al.}\cite{Hung2018APM1} used automated performance metrics (APMs) in order to predict a patient's post-operative length of stay within a hospital. Another line of research has instead focused on exclusively exploiting live surgical videos from endoscopic cameras to classify surgical activity\cite{Zia2018}, gestures\cite{Bejar2012,Luongo2021}, and skills\cite{Lavanchy2021}, among other tasks\cite{Zappella2013,Bar2020,Van2021}. Most recently, attention-based neural networks such as Transformers\cite{Vaswani2017} have been used to distinguish between distinct surgical steps within a procedure\cite{Garrow2021,Czempiel2021,Nwoye2021,Aspart2022}.

While promising, previous research exhibits two overarching limitations\cite{Lam2022}. First, the model architectures that are put forth are \textit{specialized for a single task}. Whether such models can also be used to perform other more complex tasks is an open question. Their inability to be used in such a capacity implies that entirely new models would need to be designed and evaluated, a process which is time-consuming and cumbersome for machine learning practitioners and which can delay the clinical translation of novel computational findings. Second, the models that are put forth are \textit{seldom rigorously evaluated}. For example, whether such models generalize to unseen videos from distinct surgical procedures and medical centres (with different surgeons and patient demographics) remains an open question.

In this paper, we holistically quantify robotic surgical procedures in a reliable, objective, and scalable manner via a deep learning framework, entitled Roboformer (see Fig.~\ref{fig:pipeline}). Roboformer is \textit{vision-based} in that it operates exclusively on surgical videos, which are recorded by default during robotic surgeries and thus widely accessible. Our framework is also \textit{unified} in that it is capable of independently achieving, without any modifications, multiple machine learning (ML) tasks; surgical phase recognition, gesture classification, and skills assessment (see Fig.~\ref{fig:datasets_and_tasks}c), as outlined next. 

In the context of \textit{phase recognition}, our framework received a surgical video segment and reliably identified the depicted surgical (sub)step category within the procedure. We showed that this capability extends to the setting where our framework is deployed on unseen surgical videos from a completely different medical centre. To better identify the root cause of our framework's performance, we conducted an extensive ablation study which quantified the marginal contribution of each our framework's components on its overall performance. In the context of \textit{gesture classification}, our framework identified the most likely gesture depicted in a surgical video segment. In pursuit of a rigorous evaluation setup, we demonstrated that our framework can generalize well to unseen videos, surgeons, medical centres, and surgical procedures. In the context of \textit{skills assessment}, our framework returned the skill-level at which a surgical activity was executed. In this setting, we showed that our framework can reliably distinguish between low- and high-skill surgical activity while simultaneously providing users with an explanation for its assessment. These overall findings, which were based on experiments conducted on videos of minimally-invasive robotic surgeries, point to the generalizability and potential usability of our framework in a clinical setting.

\section*{Results}

\subsection*{Data and Machine Learning Tasks}

We evaluated our framework's ability to achieve several machine learning tasks that fall within two commonly-encountered types of steps within surgery; dissection and suturing. We next outline the details of these steps and the corresponding machine learning tasks that we set out to achieve.


\subsubsection*{Dissection Step}
Dissection is a fundamental element of almost any surgery. For example, a robot-assisted radical prostatectomy procedure, in which the prostate gland is removed from a patient's body due to the presence of cancerous tissue, can entail a nerve-sparing (NS) dissection step (Fig.~\ref{fig:datasets_and_tasks}b). NS, which is one of several dissection steps in the RARP procedure, involves preserving the (left and right) neuro-vascular bundle, a mesh of vasculature and nerves surrounding the prostate, and is essential for post-operative recovery of erectile function. Similarly, a robot-assisted partial nephrectomy (RAPN), in which part of the kidney is removed from a patient's body due to the presence of cancerous tissue, entails a renal hilar dissection step (Fig.~\ref{fig:datasets_and_tasks}b). This involves removing the connective tissue around the renal artery and vein to control any potential bleeding from these blood vessels.

Although these dissection steps are procedure-specific, they can be performed with a \textit{common} vocabulary of discrete dissection actions, or gestures. In fact, there is an established dissection gesture taxonomy\cite{Ma2021} which can be used to annotate such steps. We include such annotation details in the Methods section. 

\paragraph{RARP - USC} We collected $86$ surgical videos of the NS step (both left and right) in the RARP procedure from $15$ surgeons at the University of Southern California (USC). Each video ($\approx 30$ mins) consisted of video segments ($\approx 1-2$ seconds) each depicting one of $6$ dissection gestures; cold cut, hook, clip, camera move, peel, and retraction (see Fig.~\ref{fig:datasets_and_tasks}a left column). With this, our machine learning task of interest was to identify multiple surgical gestures, also known as multi-class gesture classification, based on live surgical videos. 

\paragraph{RARP - St. Antonius Hospital} We also collected $60$ surgical videos of the RARP procedure from $8$ surgeons at St. Antonius-Hospital Gronau. Each video segment ($n=540$) was annotated with one of $6$ dissection gestures. Access to such data from a distinct medical centre and geographical location, and which are collected from different surgeons will allow us to more rigorously evaluate the generalizability of our framework.

\paragraph{RAPN - USC} We collected $27$ surgical videos of the RAPN procedure from $16$ surgeons at USC. Each video segment ($n=339$) was annotated with one of $5$ dissection gestures (based on the established gesture taxonomy\cite{Ma2021}). For this renal hilar dissection step, Hung \textit{et al.}\cite{Ma2021} previously demonstrated that the efficiency with which these gestures are performed can delineate between experienced and novice surgeons. In this setting, our machine learning task of interest remains multi-class gesture classification.

\subsubsection*{Suturing Step}
In addition to dissection, suturing is a fundamental element of surgery\cite{Moy1992}. For example, during a RARP procedure, and after the removal of the prostate gland, the bladder neck and urethra must be connected and stitched together in order to allow for the normal flow of urine. This step, which is referred to as vesico-urethral anastomosis (VUA) (Fig.~\ref{fig:datasets_and_tasks}b), is critical for ensuring that a patient does not experience internal urine leakage at the surgical site post-operatively. The VUA step typically consists of an average of 24 stitches where each stitch can be performed with a vocabulary of discrete suturing actions\cite{Luongo2021}, or gestures (see Fig.~\ref{fig:datasets_and_tasks}a right column). Note that while these gestures are common to suturing more generally, they are different to, and more subtle than, those observed with dissection. 

Each stitch can also be deconstructed into the three sub-phases of (1) needle handling, where the needle is held in preparation for the stitch, (2) needle driving, where the needle is driven through the tissue (either bladder neck or urethra), and (3) needle withdrawal, where the needle is withdrawn from the tissue. These sub-phases can also be evaluated for how well they are executed by surgeons\cite{}.

\begin{figure*}[!t]
    \centering
    \begin{subfigure}{\textwidth}
    \includegraphics[width=1\textwidth]{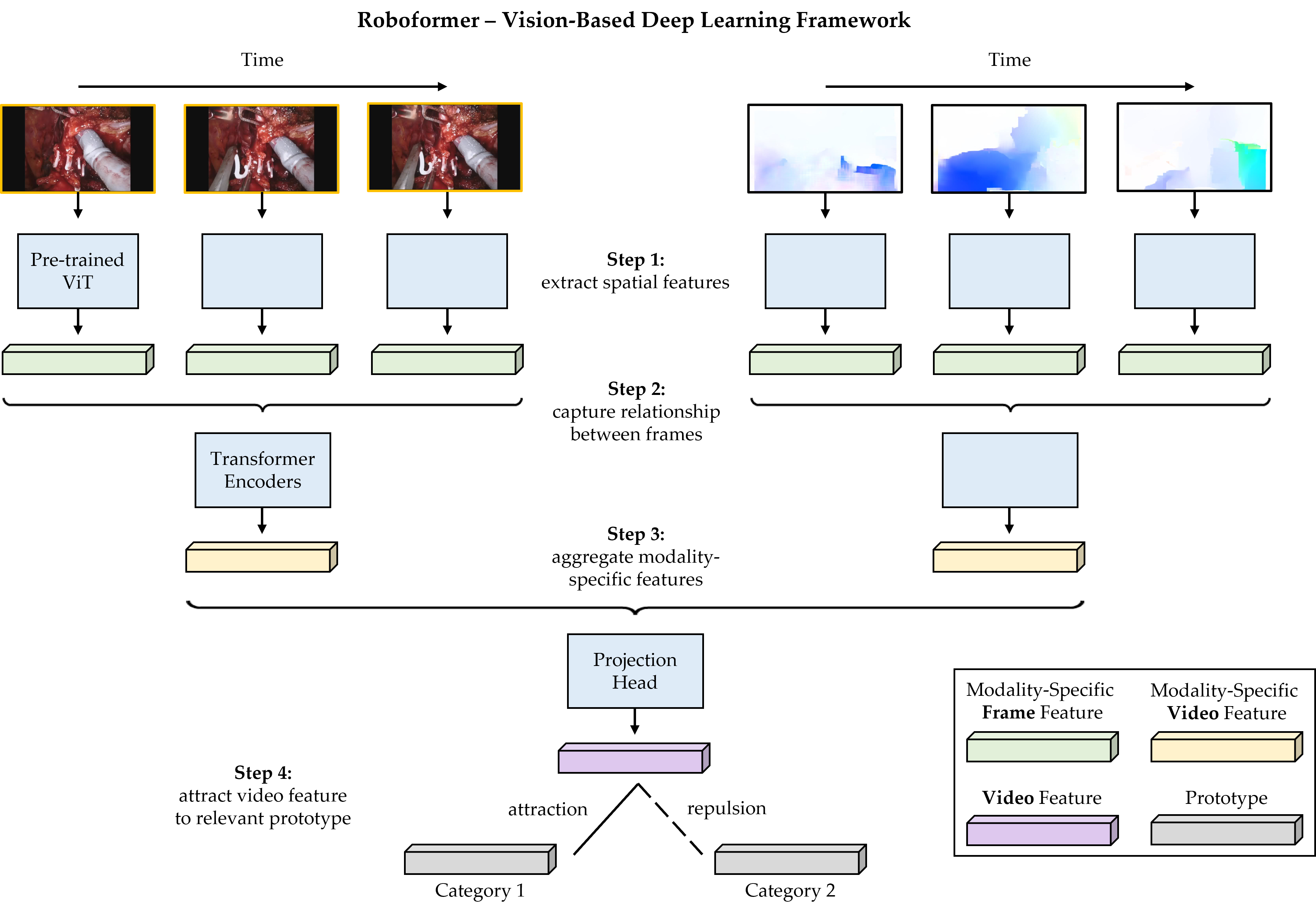}
    \end{subfigure}
    \caption{\small \textbf{Overview of the unified deep learning framework, Roboformer.} The framework consists of two parallel streams which process distinct input data modalities: RGB surgical videos and optical flow. Irrespective of the data modality, features are extracted from each frame via a Vision Transformer (ViT) pre-trained in a self-supervised manner on ImageNet. Features of video frames are then input into a Transformer encoder to obtain a modality-specific video feature. These modality-specific features are aggregated and passed into a projection head to obtain a single video feature, which is either attracted to, or repelled from, the relevant prototype, depending on the task at hand. Although we illustrate two prototypes to reflect binary outcomes (e.g., high-skill vs. low-skill activity), we would have $C$ prototypes in a setting with $C$ categories.}
    \label{fig:pipeline}
\end{figure*}

\paragraph{RARP - USC} We collected $78$ surgical videos of the VUA phase from $19$ surgeons at USC. Each video segment was annotated with the following: \textit{Sub-phases} - needle handling, needle driving, and needle withdrawal. \textit{Gestures} - right forehand under (R1), right forehand over (R2), left forehand under (L1), and combined forehand over (C1). These suturing gestures are more subtle, and thus more difficult to distinguish, than those observed for dissection since they involve fewer distinctive visual cues (e.g., different tools). \textit{Skills} - binary skills assessment labels (low-skill vs. high-skill) for the needle handling and needle driving sub-phases\cite{Sanford2022}. With this, our machine learning tasks of interest were to perform (a) multi-class surgical phase recognition, (b) multi-class gesture classification, and (c) binary skills assessment based on live surgical videos.

\paragraph{RARP - St. Antonius Hospital} We exploited the same $60$ surgical videos (mentioned above) of the RARP procedure from St. Antonius-Hospital Gronau. Each video segment in the VUA step was annotated with a sub-phase ($n=2115$) and a skills assessment label (needle handling: $n=240$, needle driving: $n=280$). The motivation for doing so, as expressed earlier, is that such data would allow us to determine the ability of our model to generalize across medical centres.

\subsection*{Algorithm Development}
Our framework, Roboformer, involved a Transformer-based architecture which consists of parallel streams that process distinct data modalities; RGB surgical videos and optical flow (see Fig.~\ref{fig:pipeline}). We extract frame-specific features using a Vision Transformer (ViT) pre-trained in a self-supervised manner on ImageNet\cite{Caron2021}. We capture the relationship between, and temporal ordering of, frames by inputting the frame features into a Transformer encoder (with a self-attention mechanism) to obtain a modality-specific video feature. By aggregating and non-linearly transforming modality-specific video features, we obtain a single video feature. This feature is either attracted to, or repelled from, prototypes which reflect distinct categories (e.g., six prototypes for six distinct gestures). A classification is made by identifying the prototype which is most similar to the video-level feature. During inference, we implement test-time augmentation (TTA), which involves exposing the network to inputs with different temporal offsets (e.g., 3 frames), in order to capture more information in the overall video which may be relevant to achieving the task. An in-depth description of the data modalities and framework is provided in the Methods section. 

\subsection*{Surgical Phase Recognition}


We first tasked our framework to distinguish between surgical sub-phases (i.e., addressing the \textit{what} of surgery). We trained our framework on the training set of the VUA step of the RARP procedure. and evaluated it using 10-fold Monte Carlo cross-validation where each fold's test set consisted of gestures from videos unseen during training. This allowed us to evaluate the framework's ability to generalize to unseen videos (hereon referred to as \textit{across videos}). A detailed breakdown of the number of instances used for training, validation, and testing can be found in Supplementary Note 1.

\begin{figure*}[!t]
    \centering
    \begin{subfigure}{1\textwidth}
    \centering
    \includegraphics[width=\textwidth]{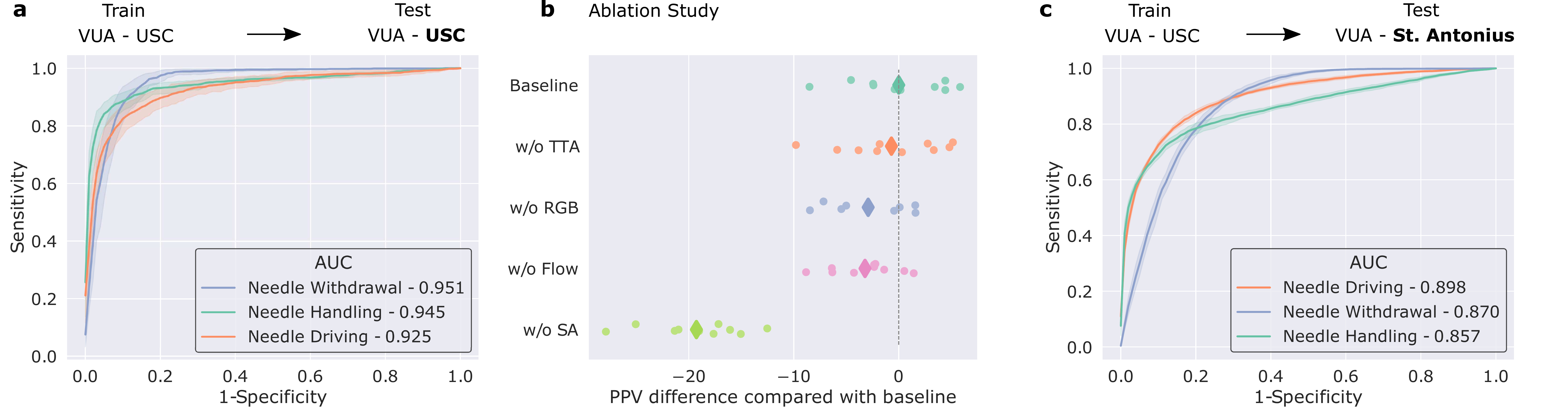}
    \end{subfigure}
    \caption{\small \textbf{Performance of the phase recognition system and marginal benefit of its components via an ablation study.} The network is trained on the USC vesico-urethral anastomosis (VUA) training set and evaluated on the \textbf{(a)} USC VUA test set and \textbf{(c)} St. Antonius VUA dataset. Results are shown as an average ($\pm 1$ standard deviation) of 10 Monte-Carlo cross-validation steps. We showed that our framework can generalize across surgical videos, surgeons, and medical centres. \textbf{(b)} We trained our framework in distinct settings to quantify the marginal benefit of its core components (see Algorithm Development) on overall performance (positive predictive value or PPV). We removed test-time augmentation (w/o TTA), RGB frames (w/o RGB), Flow frames (w/o Flow), and the self-attention mechanism (w/o SA). We show that a dual-modality input (RGB and Flow) and a self-attention mechanism are the greatest contributors to PPV.}
    \label{fig:vua_phase_recognition}
\end{figure*}

\paragraph{Across Videos} We deployed Roboformer on the test set of the VUA step of the RARP videos collected from USC, and in Fig.~\ref{fig:vua_phase_recognition}a, we present the receiver operating characteristic (ROC) curves stratified according to the various sub-phases: needle handling, needle driving, and needle withdrawal. We observed that our framework reliably distinguished between the surgical sub-phases, as evident by $\uparrow \mathrm{AUC}$ it achieved. For example, for needle driving and withdrawal, $\mathrm{AUC} = 0.925$ and $0.951$, respectively. 

\paragraph{Across Medical Centres} In order to determine whether our framework can generalize to unseen videos from a distinct medical centre, we deployed it on video segments ($n=2116$) collected from St. Antonius Hospital, Gronau, Germany (see Fig.~\ref{fig:vua_phase_recognition}c). We found that our framework continued to perform well, as evident by $\mathrm{AUC} = 0.857$, on the lower end, for needle handling and $\mathrm{AUC} = 0.898$, on the upper end, for needle driving. 

These findings bode well for the setting in which our framework is deployed on unseen videos without ground-truth sub-phase labels with the purpose of temporally segmenting these sub-phases during surgery. Recall that a single stitch consists of one of each of needle handling, driving, and withdrawal, in that order. Therefore, from a clinical perspective, the utility of segmenting these sub-phases is threefold. First, it can be used to count the total number of stitches that are performed in the VUA step, shedding light, perhaps, on surgeon-specific preferences (e.g., how far apart each stitch is made). Second, this approach can help track the duration of each stitch, thereby identifying stitches that took longer than normal to complete due to, for example, unforeseen complications. Third, and most importantly, identifying \textit{what} is happening in the surgery (e.g., needle driving), is a pre-requisite for quantifying \textit{how} these sub-phases are executed (e.g., technical skill assessment), results for which we present in subsequent sections.

\subsubsection*{Ablation Study}
We wanted to better understand the degree to which the components of our framework (see Algorithm Development) contributed to its overall performance. As such, we trained variations of the same framework, after having removed or modified these components, and evaluated them on the task of surgical phase recognition. In Fig.~\ref{fig:vua_phase_recognition}b, we present the positive predictive value (PPV) of models which are trained using our proposed framework (Baseline), evaluated without test-time augmentation (w/o TTA), and exposed to only optical flow (w/o RGB) or RGB frames (w/o Flow) as inputs. We also removed the self-attention component which captured the relationship between, and temporal ordering of, frames (w/o SA). In this setting, we simply averaged the frame features. Although we present the PPV here, we arrived at similar results when using other metrics. We observed that test-time augmentation provides a marginal benefit to the performance of our framework. This is indicated by the $\Delta \mathrm{PPV} \approx -1$, on average, of the w/o TTA setting relative to the Baseline implementation. We also observed that the dual-modality input (RGB and Flow) has a greater contribution to performance than using either modality of data alone. By removing RGB frames (w/o RGB) or optical flow (w/o Flow), the framework exhibited an average $\Delta \mathrm{PPV} \approx -3$ relative to the Baseline implementation. Such a finding suggests that these two modalities are complementary to one another. We found that the self-attention mechanism was the largest contributor to model performance, where its absence resulted in $\Delta \mathrm{PPV} \approx -20$. This finding implies that capturing the relationship between, and temporal ordering of, frames is critical to the quantification of the elements of surgery. Combined, these findings motivated our use of the Baseline model for all subsequent experiments.

\subsection*{Surgical Gesture Classification}


In the previous section, we demonstrated our framework's ability to distinguish between surgical phases (the \textit{what} of surgery) and generalize to unseen videos, and also quantified the marginal benefit of its components via an ablation study. In this section, we examine our framework's ability to distinguish between surgical gestures (the \textit{how} of surgery) in both suturing and dissection steps. We trained our framework to distinguish between four distinct gestures (R1, R2, L1, and C1) in the suturing step and six distinct gestures (c, h, k, m, p, and r) in the dissection step. This was performed on the training sets of the VUA and NS steps of the RARP procedures (from USC), respectively. We continued to evaluate our framework using 10-fold Monte Carlo cross-validation. To reiterate, this approach allowed us to evaluate our framework's ability to generalize to unseen videos, a feat which is more challenging and representative of real-world deployment than generalizing to unseen instances within the \textit{same} video.

\begin{figure*}[!t]
    \centering
    ~
    \begin{subfigure}{1\textwidth}
    \centering
    \includegraphics[width=\textwidth]{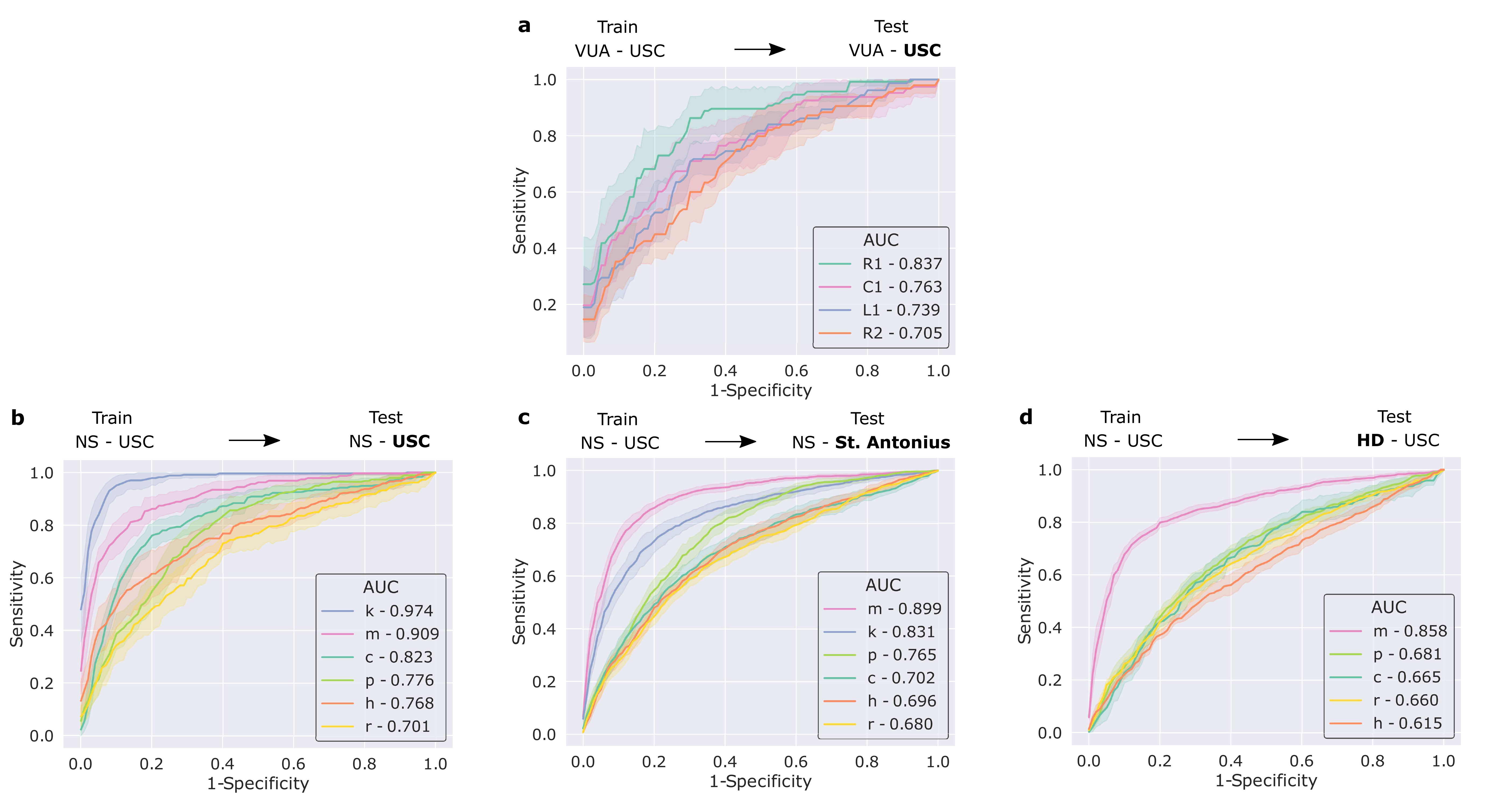}
    \end{subfigure}
    \caption{\small \textbf{Receiver operating characteristic curves of the gesture classification system.} The network is trained on the \textbf{(a)} USC vesico-urethral anastomosis (VUA) training set and evaluated on the USC VUA test set. The gestures are right forehand under (R1), right forehand over (R2), left forehand under (L1), and combined forehand over (C1). In \textbf{(b)} - \textbf{(d)}, the network is trained on the USC nerve-sparing (NS) training set. It is evaluated, however, on the \textbf{(b)} USC NS test set, \textbf{(c)} St. Antonius NS set, and \textbf{(d)} USC renal hilar dissection (HD) set. The gestures are cold cut (c), hook (h), clip (k), camera move (m), peel (p), retraction (r). Note that clips (k) are \textit{not} used during the renal hilar dissection step. Results are shown as an average ($\pm 1$ standard deviation) of 10 Monte-Carlo cross-validation steps. We show that the network can generalize to unseen videos, surgeons, medical centres, and surgical procedures.}
    \label{fig:gesture_classification_roc_curves}
\end{figure*}

\paragraph{Across Videos} For the suturing and dissection steps, we deployed the framework on the test set of VUA and NS data collected from USC, respectively. The results of those experiments are depicted in Fig.~\ref{fig:gesture_classification_roc_curves}a and b, respectively, where we present the ROC curves of the corresponding predictions stratified according to the gestures. There are two main takeaways here. First, we observed that the framework can generalize well to both suturing and dissection gestures in unseen videos. This is exhibited by $\uparrow \mathrm{AUC}$ achieved by the framework across the gestures. For example, in the suturing step, $\mathrm{AUC} = 0.837$ and $0.763$ for the right forehand under (R1) and combined forehand over (C1) gestures. In the dissection step, $\mathrm{AUC} = 0.974$ and $0.909$ for the clip (k) and camera move (m) gestures. These findings bode well for the potential deployment of our framework on unseen videos for which ground-truth gesture labels are unavailable, an avenue we explore in a subsequent section. Second, we found that the performance of the framework differs across the gestures. For example, in the dissection step, it achieved $\mathrm{AUC} = 0.701$ for the retraction (r) gesture and $\mathrm{AUC} = 0.974$ for the clip (k) gesture. We hypothesize that the strong performance of the framework for the latter stems from the clear visual presence of a clip in the surgical field of view. In contrast, we believe that the relatively lower ability to identify retractions is driven by its ubiquitous presence in the surgical field of view, as explained next. Retraction is often performed by, and annotated as such when, the dominant hand is \textit{actively} performing this gesture. However, retraction, which is a core gesture that is used to, for example, improve visualization of the surgical field, often \textit{complements} other gestures. As such, it appears alongside other gestures and can thus be confused for them by the model. 

\paragraph{Across Medical Centres} In addition to evaluating our framework's ability to generalize to unseen videos, we explored the degree to which it can generalize to videos from a distinct medical centre. To that end, we deployed our framework on nerve-sparing dissection data of RARP procedures collected from St. Antonius Hospital, Gronau, Germany (see Fig.~\ref{fig:gesture_classification_roc_curves}c). We found that our framework continues to perform well in such a setting despite the data being collected from unseen surgeons operating on distinct patients at a medical centre in a different geographical location. For example, it achieved $\mathrm{AUC} = 0.899$ and $0.831$ for the gestures of camera move (m) and clip (k), respectively. Such a finding suggests that our framework can also be reliably deployed on data with multiple sources of variability (surgeon, location, etc.). We expected, and indeed observed, a slight degradation in performance in this setting relative to when the framework was deployed on data collected from USC. For example, $\mathrm{AUC} = 0.823 \rightarrow 0.702$ for the cold cut (c) gesture in the USC and St. Antonius settings, respectively. This was expected due to the potential shift in the distribution of data collected across the two medical centres, which has been documented to negatively affect network performance\cite{Kiyasseh2021}. Potential sources of distribution shift include variability in how surgeons perform the same set of gestures (e.g., different techniques) and the camera recording devices between the surgical robots. Furthermore, our hypothesis for why this degradation affects certain gestures (e.g., cold cuts) more than others (e.g., clips) is that the latter exhibits less variability than the former, and is thus easier to classify by the model. 

\begin{figure*}[!b]
    \centering
    ~
    \begin{subfigure}{1\textwidth}
    \centering
    \includegraphics[width=1\textwidth]{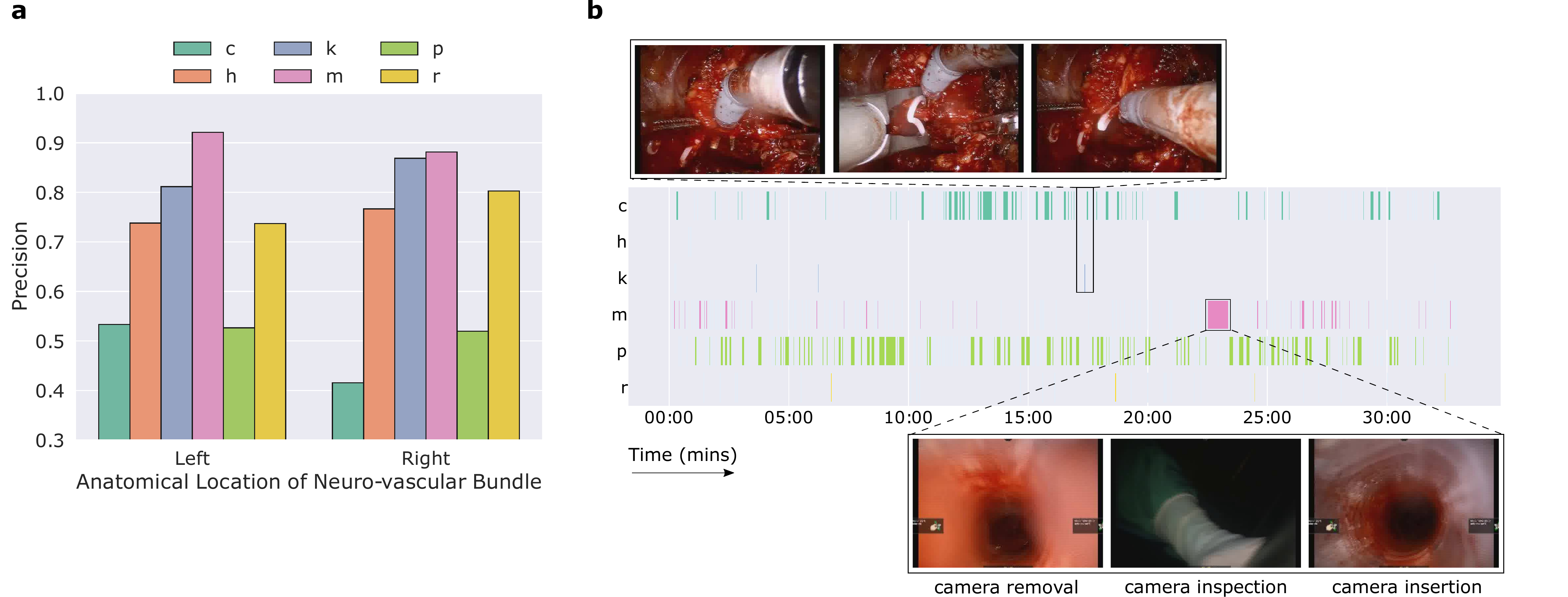}
    \end{subfigure}
    \caption{\small \textbf{Quantitative and qualitative evaluation of the gesture classification framework when deployed on nerve-sparing videos without ground-truth gesture labels.} \textbf{(a)} Proportion of predicted gestures identified as correct (precision) stratified based on the anatomical location of the neuro-vascular bundle in which the gesture are performed. \textbf{(b)} Each row represents a distinct gesture and each vertical line represents the occurrence of that gesture over time. Our framework identified a sequence of gestures (hook, clip, cold cut) that is typical in the nerve-sparing step of RARP procedures. Furthermore, our framework was able to identify outlier behaviour as reflected by a longer-than-normal camera move gesture which coincided with the removal, inspection, and re-insertion of the camera into the patient's body.}
    \label{fig:gesture_classification_unlabelled}
\end{figure*}

\paragraph{Across Surgical Procedures} Having shown that our framework can generalize to unseen videos, surgeons, and medical centres, we also wanted to explore the degree to which our framework can generalize to unseen \textit{surgical procedures}. Dissection is a common step in many surgical procedures, and we therefore expect to see similar dissection gestures across such procedures. Despite this commonality, we expected this setting to be even more challenging than those presented earlier due to, for example, distinct (and potentially distracting) anatomical landmarks between surgical procedures. To that end, we deployed the framework on the renal hilar dissection (HD) step of the robot-assisted partial nephrectomy (RAPN) procedure (see Fig.~\ref{fig:gesture_classification_roc_curves}d). Note that the clips (k) are not used in this step, and we therefore do not report its performance. We observed that our gesture classification framework manages to adequately generalize to unseen surgical procedures, albeit exhibiting degraded performance, as expected. Specifically, it achieved $0.615 < \mathrm{AUC} < 0.858$ across the gestures. Interestingly, the hook (h) gesture experienced the largest degradation in performance (h) ($\mathrm{AUC} = 0.768 \rightarrow 0.615$). To explain this, we hypothesized that this was due to the difference in tissue in which a hook is performed. Whereas in the nerve-sparing dissection step, a hook is typically performed in the neuro-vascular bundle (a region of nerves and vessels), in the renal hilar dissection step, it is performed in the connective tissue around the renal artery and vein, delivering blood to and from the kidney, respectively.  

\subsubsection*{Inference on Data Without Gesture Labels}
The ultimate goal of an automated gesture classification system is to deploy it on surgical videos \textit{without} any ground-truth gesture labels and identify the intra-operative sequence of gestures performed by a surgeon. 

In pursuit of that goal, we deployed our framework, having been trained on the USC NS training set, on a subset of the USC NS videos ($n=154$) without any gesture labels. Specifically, we segmented the entire surgical step (on the order of 30 minutes) into 1-second non-overlapping intervals and deployed our model to output a probability distribution reflecting how likely it is a for a particular gesture to be present in each interval. In the process, we wanted to account for the event that the surgical field of view may depict activity or objects that the model had not seen during training. This can include, for example, less common surgeon-specific gestures or unexpected behaviour. Expressed differently, we wanted to be highly confident in the gesture predictions being made while flagging potential outlier video segments. To achieve this, we took inspiration from the work on neural network ensembles \cite{Lakshminarayanan2017} and out-of-distribution (OOD) detection \cite{Roy2022}. In that body of work, data points whose model predictions reflect a high degree of uncertainty, as quantified by a measure of entropy, are deemed out-of-distribution. Therefore, instead of deploying a single model per 1-second interval, we deployed all 10 of our models (which were trained on slightly different subsets of data). By averaging the output probabilities across the models, we made a gesture prediction by identifying the most probable gesture. If the uncertainty (entropy) of the averaged probabilities exceeded some threshold, the model abstained from making a prediction. In doing so, we aimed to avoid potentially erroneous gesture predictions. Further details can be found in the Methods section. 

We evaluated our framework, when deployed on data without ground-truth gesture labels, both quantitatively and qualitatively. 

\paragraph{Quantitative Evaluation} We first randomly selected one prediction per gesture category in each video, for a total of $800$ gesture predictions. Doing so ensured we retrieved predictions from a representative and diverse set of videos, thus improving the generalizability of our findings. We then manually inspected the corresponding video segments to confirm whether or not this prediction was correct. In Fig.~\ref{fig:gesture_classification_unlabelled}a, we report the precision of these gesture predictions (proportion of predicted gestures which are correct). To gain a better understanding of whether gestures are more easily identified when performed in distinct anatomical locations, we also stratify the results based on the anatomical location of the neuro-vascular bundle (left vs. right). These two locations are the primary anatomical landmarks within the nerve-sparing dissection step. We observed that the precision of the predictions ranged from $0.40$ (right cold cut) to $0.90$ (left camera move). After additional inspection, we discovered that the cold cut (c) gesture predictions were still identifying a cutting gesture, albeit of a different type, known as a \textit{hot cut} which involves applying heat/energy to cut tissue. We also found that the anatomical location in which a gesture is performed has a minimal effect on the gesture classification framework. This is evident by the similar performance achieved in the left and right neuro-vascular bundles. For example, hook (h) gesture predictions exhibited precision of $\approx 0.75$ in both locations. Such a finding suggests that our framework can be reliably deployed on surgical videos of the nerve-sparing step. 

\paragraph{Qualitative Evaluation} To qualitatively evaluate the performance of our framework, we present its gesture predictions for a single $30$-minute nerve-sparing video (see Fig.~\ref{fig:gesture_classification_unlabelled}b). Each row represents a distinct gesture and each vertical line represents the occurrence of this gesture over time. We observed that, although the model was not explicitly informed about the relationship between gestures, it nonetheless correctly identified a \textit{sequence} (ordering) of gestures which is typical of the nerve-sparing step within RARP procedures. The sequence, which involves a 1) hook, 2) clip, and 3) cold cut, is performed to separate the neuro-vascular bundle from the prostate while minimizing the degree of bleeding that the patient incurs. Such a finding qualitatively reaffirms our earlier claims that our framework is capable of correctly classifying surgical gestures. 

We also found that our framework is capable of identifying outlier behaviour, despite not being explicitly trained to do so, as explained next. Broadly speaking, the camera move (m) gesture typically entails a brief shift (on the order of 1 second) in the surgical field of view. However, our framework identified a 1-minute interval ($60\times$ longer than normal) during which a camera move is performed. Under normal circumstances, this would be considered aberrant behaviour. Indeed, after manual inspection, we discovered that this time interval coincided with the removal of the camera from the patient's body, its inspection, and its re-insertion into the patient's body. Beyond identifying outliers, this capability also allows for the exclusion of irrelevant frames for, and thus reducing the amount of noise injected into, downstream video analysis. 

\subsection*{Surgical Skills Assessment}


At this point, we have demonstrated that the same unified framework can independently achieve the two tasks of surgical phase recognition (the \textit{what} of surgery) and gesture classification (the \textit{how} of surgery), and generalize to unseen videos in the process. In this section, we examine our framework's ability to distinguish between low- and high-skill activities performed by surgeons. In doing so, we also address the \textit{how} of surgery however through the lens of surgeon technical skill. 

We evaluated the quality with which two suturing sub-phases were executed by surgeons. These two sub-phases are needle handling and needle driving (see Fig.~\ref{fig:datasets_and_tasks}a right column). For each sub-phase, we trained our framework to distinguish between low- and high-skill activity. For needle handling, a high-skill assessment is based on the number of times the surgeon must reposition their grip on the needle in preparation for driving it through the tissue (the fewer the better). For needle driving, a high-skill assessment is based on the smoothness and number of adjustments required to drive the needle through the tissue (the smoother and fewer number of adjustments the better). We continued to evaluate our framework using 10-fold Monte Carlo cross-validation. This allowed us to evaluate the framework's ability to generalize to needle handling and driving in unseen videos. 

\begin{figure*}[!t]
    \centering
    \begin{subfigure}{1\textwidth}
    \centering
    \includegraphics[width=1\textwidth]{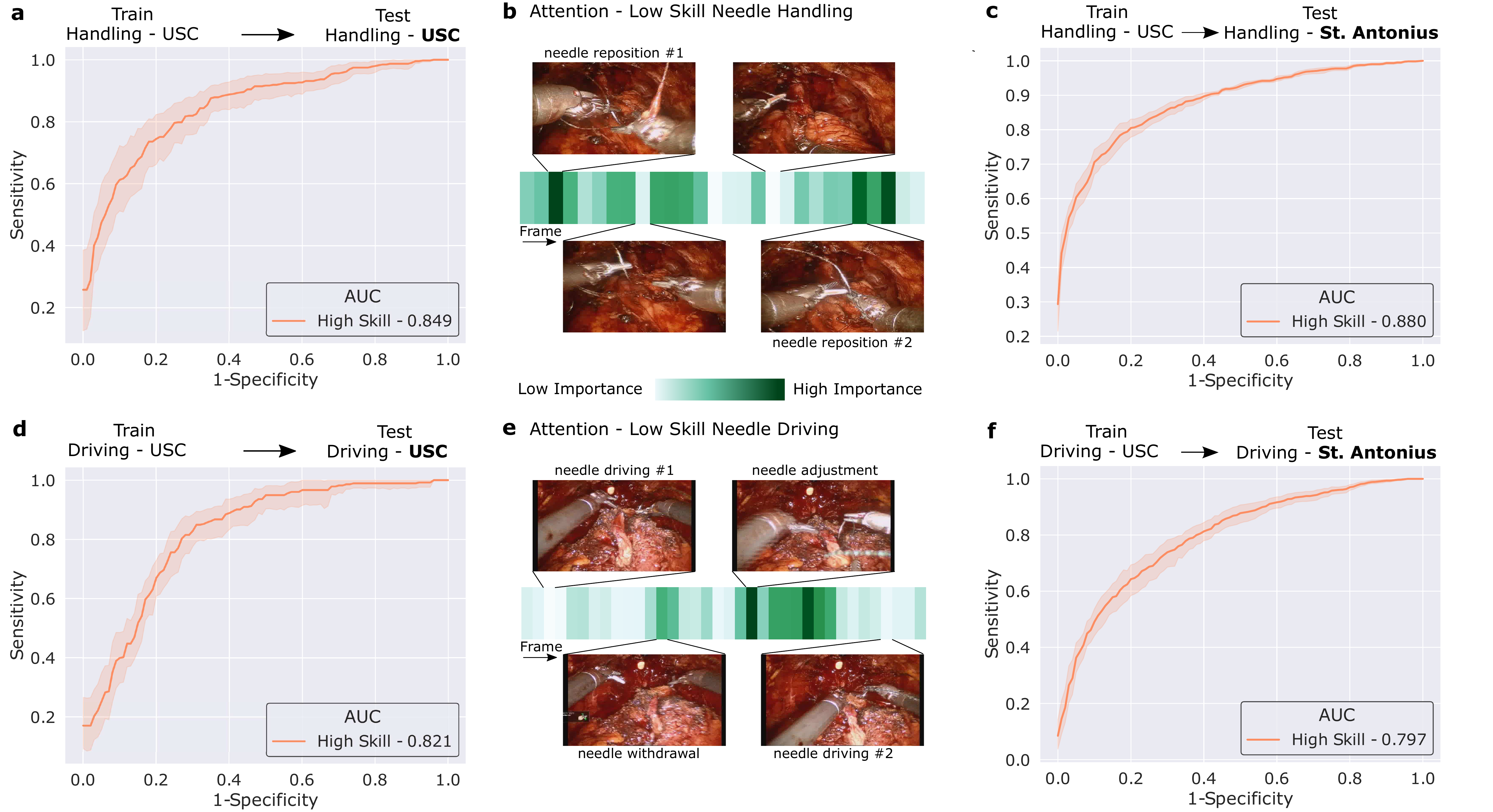}
    \end{subfigure}
    \caption{\small \textbf{Receiver operating characteristic curves of the skills assessment system and temporal attention across visual frames.} \textbf{(Top - Needle Handling)} The network is trained on the USC training set of needle handling and evaluated on the  \textbf{(a)} USC test set and \textbf{(c)} St. Antonius dataset. \textbf{(Bottom - Needle Driving)} The network is trained on the USC training set of needle driving and evaluated on the  \textbf{(d)} USC test set and \textbf{(f)} St. Antonius dataset. Results are shown as an average ($\pm 1$ standard deviation) of 10 Monte-Carlo cross-validation steps. We show that our framework can generalize across videos, surgeons, and medical centres. In \textbf{(b)} and \textbf{(e)}, we depict the temporal attention (darker shade is higher attention) placed on frames of a video segment depicting low skill needle handling and needle driving, respectively. In \textbf{(b)}, we see that the frames with high attention depict needle repositions, which is in alignment with the low-skill category of needle handling. In \textbf{(e)}, we see that the frames with high attention depict needle adjustments (e.g. needle withdrawal in the opposite direction of motion), which is in alignment with the low-skill category of needle driving.}
    \label{fig:vua_skills_assessment_roc_curves}
\end{figure*}

\paragraph{Across Videos} We deployed our framework on the test set of the VUA step of the RARP videos collected from USC, and in Fig.~\ref{fig:vua_skills_assessment_roc_curves}a and d, we present the ROC curves associated with the skill of needle handling and driving, respectively. We found that our framework can reliably distinguish between low- and high-skill surgical activity, achieving $\mathrm{AUC} = 0.849$ and $0.821$ for the needle handling and driving activity, respectively. These results indicate that our framework was able to pick up on a distinctive pattern in the data associated with the quality of the skill being assessed.

\paragraph{Across Medical Centres}
As in previous sections, we deployed our framework on unseen video segments collected from St. Antonius Hospital, Gronau, Germany (see Fig.~\ref{fig:vua_skills_assessment_roc_curves}c and f). We found that our framework continued to reliably distinguish between low- and high-skill activity ($\mathrm{AUC}=0.880$ and  $0.797$ for needle handling and needle driving, respectively). This relatively strong performance is achieved despite the framework being exposed to data from a different medical centre whereby surgeon preferences (and perhaps patient demographics) are likely to introduce variability in the way activities are executed. Such a finding also lends support to the objective nature of the ground-truth skills assessment annotations. 


\paragraph{Explainability of Skills Assessment} Beyond assessing the performance of our framework, we also wanted to confirm whether or not the framework was identifying clinically-relevant visual cues while automating skills assessment, an important aspect of clinical AI frameworks\cite{Ghassemi2021}. Doing so would instill machine learning practitioners with confidence that the framework is indeed latching onto appropriate features, and can thus be trusted in the event of future deployment within a clinical setting. We quantified this by first sampling a video instance depicting a low-skill activity (needle handling or needle driving) which was correctly classified by the framework. We then retrieved the temporal attention that was placed by the Transformer encoder (see Fig.~\ref{fig:pipeline} for network architecture) on each of the frames in this video segment. In an ideal situation, high attention would be placed on frames of relevance, where relevance is clinically-defined and specific to the skill of interest. 

In Fig.~\ref{fig:vua_skills_assessment_roc_curves}b and e, we present the attention placed on frames within a needle handling and needle driving video segment, respectively, each of which is classified as low skill. The darker the shade, the greater the attention. We observed that, in both settings, the framework places attention on frames which reflect clinically-relevant visual states. For example, in Fig.~\ref{fig:vua_skills_assessment_roc_curves}b, we observed that the frames with a high degree of attention depict a situation where both instruments are simultaneously holding onto the needle. This manoeuvre is typically performed when the surgeon needs to hold the needle steady with one instrument while repositioning the other instrument's grasp of the needle, and, as such, multiple repetitions of this behaviour aligns well with low-skill category of needle handling. For needle driving (Fig.~\ref{fig:vua_skills_assessment_roc_curves}e), we found that the video segment depicted a needle which had initially been driven through the tissue, required adjustment and was thus completely withdrawn (opposite to direction of motion) before being re-driven through the tissue seconds later. The framework placed a high level of attention on the withdrawal of the needle and its adjustment, in alignment with the low-skill category of needle driving. More broadly, these explainable findings suggest that our framework is not only capable of providing surgeons with a scalable assessment of skill but can also pinpoint the problematic scenes in the video segment. This capability, which in some sense addresses \textit{why} an assessment of low skill was made, bodes well in the event such a framework is deployed to provide surgeons with \textit{targeted} feedback. 

\subsection*{Combining the elements of surgery}
Thus far, we have presented our machine learning tasks, which quantify the various elements of surgery, as independent of one another. Considering these tasks in unison, however, suggests that our framework can provide a surgeon with reliable, objective, and scalable feedback of the following form: \enquote{when completing stitch number 3 of the suturing step, your needle handling (what - \textit{sub-phase}) was executed poorly (how - \textit{skill}). This is likely due to your activity in the first and final quarters of the needle handling sub-phase (why - \textit{attention})}. Such granular and temporally-localized feedback allows a surgeon to better focus on the element of surgery that requires improvement. As such, a surgeon can now better identify, and learn to avoid, problematic intra-operative surgical behaviour in the future.

\section*{Discussion}
In this paper, we aimed to reliably automate the holistic quantification of surgery. To do so, we presented a unified deep learning framework, Roboformer, capable of independently achieving multiple tasks in the surgical domain. These tasks ranged from surgical phase recognition (the \textit{what} of surgery) to gesture classification and skills assessment (the \textit{how} of surgery). In the context of surgical phase recognition, we showed that our framework can generalize well across videos, surgeons, and medical centres. We observed that test-time augmentation and dual-modality inputs, in the form of RGB frames and optical flow, contribute to the strong performance of our framework. In the context of gesture classification, we also demonstrated that our framework was able to generalize across videos, surgeons, and medical centres. We also showed that our framework can generalize to videos of a completely different surgical procedure. In the context of skills assessment, our framework was capable of distinguishing between low- and high-skill surgical activity, even when deployed on videos collected from a distinct medical centre. While performing this skills assessment, our framework simultaneously identified relevant frames in the surgical video for such an assessment.

Our findings provide the initial evidence in support of the eventual use of deep learning frameworks in the domain of surgery. To begin, the ability of our framework to generalize to videos across a wide range of settings (e.g., medical centres and surgical procedures) should instill surgeons with confidence as it pertains to its clinical deployment. Furthermore, our framework's ability to provide explainable results, by pinpointing temporal frames which it deems relevant to its prediction, has a threefold benefit. First, explainability can improve the trustworthiness of an AI framework, a critical pre-requisite for clinical adoption\cite{Lam2022}. Second, it allows for the provision of more targeted feedback to surgeons about their performance. This, in turn, would allow them to modulate such performance as a means to improve patient outcomes. Third, an element of explainability implies improved framework transparency and is viewed, by some, as contributing to the ethical deployment of AI models in a clinical setting\cite{Lam2021}.  

Our framework, however, does suffer from several limitations. We opted to extract features from frames via a vision transformer (with frozen parameters) that was pre-trained on ImageNet. Although this significantly reduced the computational burden associated with training and deploying our framework, it may have limited the expressiveness of the extracted representations. This is because of the domain shift that exists between data in ImageNet and surgical video data. We leave it to future work to explore the benefit of pre-training and fine-tuning vision transformers on surgical videos.  Furthermore, after some exploration, we decided to use a strategy that selected the same subset of frames from each video segment as input to the framework. Although we found this to work for our purposes, it is likely that an optimal choice of frames exists. Identifying these relevant candidate frames a priori and dynamically selecting them during training and deployment can yield improved performance. Lastly, we acknowledge that skills assessment can be a subjective matter which is confounded by factors such as surgeon preference and variable patient anatomy. This subjectivity can bias the annotation of skills assessment despite the seemingly objective criteria that guide annotators. As such, label noise, in this context, may continue to pose a challenge to both training and reliably evaluating frameworks\cite{Hung2021}. 

Moving forward, we aim to leverage our framework to annotate, at scale, previously unlabelled databases of surgical videos with surgical gestures, phases, and skills assessment labels. Doing so will provide us with a more quantifiable and holistic view of surgical procedures. Equipped with this information, we see two potential avenues. First, we can start to identify correlations, and causal relationships\cite{Gowrisankaran2006}, between quantifiable surgical metrics and long-term patient outcomes. In the context of radical prostatectomy procedures, these outcomes can include the post-operative ability of a patient to have an erection (erectile function recovery) and voluntary control over their urination (urinary continence recovery). The identification of these causal relationships leads us to our second future avenue. We first look to validate these relationships in a low-stakes controlled laboratory environment before incorporating them into formal surgical training and feedback protocols. Another element we are looking to study is the degree of algorithmic bias exhibited by our proposed deep learning framework against surgeon cohorts\cite{Collins2021}. This is critical in this context of surgical feedback as a biased system can disadvantage certain surgeons (e.g., those from different age groups) and thus affect their ability to develop professionally. Overall, these considerations would ensure that the entire pipeline of surgeons from surgical residents to fellows and attendings are safely equipped to identify positive and negative behaviours in surgical procedures, can incorporate these discoveries into future surgeries, and ultimately improve patient outcomes. We hope the broader community joins us in this endeavour.

\section*{Methods}

\subsection*{Ethics Approval} All datasets (data from USC and St. Antonius Hospital) were collected under Institutional Review Board (IRB) approval in which informed consent was obtained (HS-17-00113). Moreover, the datasets were de-identifed prior to model development.

\subsection*{Data Modalities}
We designed a unified deep learning framework that is capable of achieving multiple tasks independently of one another. To capture both visual and motion cues in surgical videos, the framework operated on two distinct modalities; live surgical videos in the form of RGB frames and the corresponding optical flow of such frames. Surgical videos can be recorded at various sampling rates, which have the units of frames per second (fps). 

Knowledge of the sampling rate alongside the natural rate with which activity occurs in a surgical setting is essential to multiple decisions. These can range from the number of frames to present to a deep learning network, and the appropriate rate with which to downsample videos, to the temporal step-size used to derive optical flow maps, as outlined next. Including too many frames where there is very little change in the visual scene leads to a computational burden and may result in over-fitting due to the inclusion of highly similar frames (low visual diversity). On the other hand, including too few frames might result in missing visual information pertinent to the task at hand. Similarly, deriving reasonable optical flow maps, which is a function of a \textit{pair} of images which are temporally-spaced, is contingent upon the time that has lapsed between such images. Too short of a time-span could result in minimal motion in the visual scene, thus resulting in uninformative optical flow maps. Analogously, too long of a time-span could mean missing out on informative intermediate motion in the visual scene. We refer to these decisions as hyperparameters and outline our choices in the Hyperparameters section. Throughout this paper, we derived optical flow maps by deploying a RAFT model\cite{Teed2020}, which we found to provide reasonable maps.  

\subsection*{Roboformer Deep Learning Framework}
We extract a sequence of $D$-dimensional representations, $\{ \boldsymbol{v}_{t} \in \mathbb{R}^{D} \}_{t=1}^{T}$, from $T$ temporally-ordered frames via a (frozen) Vision Transformer (ViT) pre-trained on the ImageNet dataset in a self-supervised manner\cite{Caron2021}. Vision Transformers convert each input frame into a set of square image patches of dimension $H \times H$ and introduce a self-attention mechanism that attempts to capture the relationship between image patches (i.e., spatial information). We chose this feature extractor based on (a) recent evidence favouring self-supervised pre-trained models relative to their supervised counterparts and (b) the desire to reduce the computational burden associated with training a feature extractor in an end-to-end manner. 

We append a learnable $D$-dimensional classification embedding, $e_{\mathrm{cls}} \in \mathbb{R}^{D}$, to the beginning of the sequence of frame representations, $\{ \boldsymbol{v}_{t} \}_{t=1}^{T}$. To capture the temporal ordering of the images frames, we add $D$-dimensional temporal positional embeddings, $\{ e_{t} \in \mathbb{R}^{D} \}_{t=1}^{T}$, to the sequence of frame representations before inputting the sequence into four Transformer encoder layers. Such an encoder has a self-attention mechanism whereby each frame attends to every other frame in the sequence. As such, both short- and long-range dependencies between frames are captured. We summarize the modality-specific video through a modality-specific video representation, $h_{\mathrm{cls}} \in \mathbb{R}^{D}$, of the classification embedding, $e_{\mathrm{cls}}$, at the final layer of the Transformer encoder, as is typically done. This process is repeated for the optical flow modality stream. 

The two modality-specific video representations, $h_{\mathrm{RGB}}$ and $h_{\mathrm{Flow}}$, are aggregated as follows:

\begin{equation}
    h_{\mathrm{agg}} = h_{\mathrm{RGB}} + h_{\mathrm{Flow}}
\end{equation}

The aggregated representation, $h_{\mathrm{agg}}$, is passed through two projection heads, in the form of linear layers with a non-linear activation function (ReLU), to obtain an $E$-dimensional video representation, $h_{\mathrm{Video}} \in \mathbb{R}^{E}$. 

\subsubsection*{Training Protocol}
To achieve the task of interest, the video-specific representation, $h_{\mathrm{Video}}$, undergoes a series of attractions and repulsions with learnable embeddings, which we refer to as prototypes. Each prototype, $p$, reflects a single category of interest and is of the same dimensionality as $h_{\mathrm{Video}}$. The representation, $h_{\mathrm{Video}} \in \mathbb{R}^{E}$, of a video from a particular category, $c$, is attracted to the single prototype, $p_{c} \in \mathbb{R}^{E}$, associated with the same category, and repelled from all other prototypes, $\{ p_{j} \}_{j=1}^{C}$, $j \neq c$, where $C$ is the total number of categories. We achieve this by leveraging contrastive learning and minimizing the InfoNCE loss, $\mathcal{L}_{NCE}$. 

\begin{equation}
    \mathcal{L}_{NCE} = - \sum_{i=1}^{B} \log \frac{e^{s(h_{\mathrm{Video}},p_{c})}}{\sum_{j} e^{s(h_{\mathrm{Video}},p_{j})}}
\end{equation}

\begin{equation*}
    s(h_{\mathrm{Video}},p_{j}) =  \frac{h_{\mathrm{Video}} \cdot p_{j}}{|h_{\mathrm{Video}}||p_{j}|}
\end{equation*}

During training, we share the parameters of the Transformer encoder across modalities in order to avoid over-fitting. As such, we learn, in an end-to-end manner, the parameters of the Transformer encoder, the classification token embedding, the temporal positional embeddings, the parameters of the projection head, and the category-specific prototypes. 

\subsubsection*{Evaluation Protocol}
In order to classify a video into one of the categories, we calculate the similarity (e.g., cosine similarity) between the video representation, $h_{\mathrm{Video}}$, and each of the prototypes, $\{ p_{j} \}_{j=1}^{C}$. We apply the softmax function to these similarity values in order to obtain a probability mass function over the categories. By identifying the category with the highest probability mass (argmax), we can make a classification. 

The video representation, $h_{\mathrm{Video}}$, can be dependent on the choice of frames (both RGB and optical flow) which are initially input into the model. Therefore, to account for this dependence and avoid missing potentially informative frames during inference, we deploy what is known as test-time augmentation (TTA). This involves augmenting the same input multiple times during inference, which, in turn, outputs multiple probability mass functions. We can then average these probability mass functions, analogous to an ensemble model, to make a single classification. In our context, we used three test-time inputs; the original set of frames at a fixed sampling rate, those perturbed by offsetting the start-frame by $K$ frames at the same sampling rate. Doing so ensures that there is minimal frame overlap across the augmented inputs, thus capturing different information, while continuing to span the most relevant aspects of the video.

\subsection*{Inference on Unlabelled Data}
In the context of data without gesture labels, we performed inference on 1-second non-overlapping intervals of video segments. We chose 1-second intervals because the majority of the gestures in the training set were \textit{at least} 1-second in duration. For each 1-second interval, we performed a forward pass through each of the 10 already-trained models. This, in turn, resulted in ten probability distributions, $\{ s^{i} \in \mathbb{R}^{6} \}_{i=1}^{10}$ over the six possible gestures. By averaging these probability distributions across the models, we obtain $\bar{s} \in \mathbb{R}^{6}$ and are effectively implementing an ensemble model. In order to identify highly uncertain predictions, we calculated the entropy, $S$, of the averaged probability distribution and only considered predictions whose entropy did not exceed some threshold, $S_{\mathrm{thresh}}$ as shown below. 
\begin{equation}
    S = - \sum_{g=1}^{6} \bar{s}_{g} \log \bar{s}_{g} \leq S_{\mathrm{thresh}}
\end{equation}
\begin{equation*}
    \bar{s}_{g} = \sum_{i=1}^{10} s_{g}^{i} \quad \forall g \in [1,6]
\end{equation*}
After excluding predictions with a high level of uncertainty, we wanted to account for the fact that the same gesture can span multiple seconds, even though our prediction intervals were limited to 1 second. To do so, we made the assumption that if the same gesture were to be predicted by a model within a 2-second time interval, then it is highly likely that the entirety of that time interval reflects a single unified gesture. For example, if a retraction (r) is predicted at intervals 10-11s, 11-12s, and 15-16s, we treated this as 2 distinct retraction gestures. The first spans 2 seconds (10-12s) and the seconds spans 1 second (15-16s). This avoids us tagging spurious and incomplete gestures (e.g., the beginning or end of a gesture) as an entirely distinct gesture over time. We chose a 2-second interval to introduce some tolerance for a potential misclassification between gestures of the same type and to allow for temporal continuity of the gestures. 

\subsection*{Hyperparameters}
During training and inference, we downsample the RGB frames to $\mathrm{fps}=2$ as we found this to be a good trade-off between computational complexity and capturing sufficiently informative signals in the video to complete the task. Similarly, optical flow maps were based on pairs of images that were $0.5$ seconds apart. Shorter time-spans resulted in frames which exhibited minimal motion and thus uninformative flow maps. During training, to ensure that the RGB and optical flow maps were associated with the same time-span, we retrieved maps that overlapped in time with the RGB frames. During inference, and for test-time augmentation, we offset both RGB and optical flow frames by $K=3$ and $K=6$ frames. 

We conduct our experiments in PyTorch\cite{Paszke2019} using a V100 GPU on a DGX machine. Unless otherwise stated, we use a mini-batch size of $8$ video segments and a learning rate of $1e^{-1}$, and optimize the parameters of the model via stochastic gradient descent. Each RGB and optical flow frame was resized to $224 \times 224$ before being input into the ViT feature extractor. Input frames were pre-processed, by the ViT feature extractor, into a set of square patches of dimension $H=16$. Whereas frame representations are of dimension $D=384$, video representations and prototypes are of dimension $E=256$. 

\subsection*{Data Splits}
For all experiments, we split the data at the \textit{case video level} into a training ($90\%$) and test set ($10\%$). We used $10\%$ of the videos in the training set to form a validation set with which we performed hyperparameter tuning. By splitting at the video-level, whereby data from the same video do not appear across the sets, we are rigorously evaluating whether the model generalizes across unseen videos. This is more challenging than the setting commonly employed in previous work in which data from the same video are leaked across sets. The latter becomes trivial if the model, for example, latches onto video-specific features to achieve the machine learning task, an observation we confirmed in early experiments (not shown). 

In order to increase our confidence in the performance of our model, we evaluate it using 10-fold Monte Carlo cross-validation where possible. This involves using a different set of videos for the training, validation, and test splits outlined above, while continuing to perform the split at the case video level. The data-splits can be found in Supplementary Note 1. The performance of models is reported as an average, with a standard deviation, across the folds. 

\subsection*{Ground-Truth Annotations}
In this paper, we trained our deep learning framework in a \textit{supervised} manner by providing it with surgical video segments and ground-truth annotations of these segments. These annotations ranged from dissection or suturing gestures to sub-phases and skills assessments. In this section, we outline how we obtained such ground-truth annotations.

In all cases described below, during the training phase of the annotation process, each human rater provided with, and asked to annotate, the same subset of surgical videos. This allowed us to calculate the inter-rater reliability (between 0 and 1) of the annotations. Once this reliability exceeded $0.8$, we deemed the training phase complete.

\paragraph{Phase Annotations} A group of medical students was educated on the different sub-phases of a single stitch within a suturing step. Once the training phase of the annotation process was complete, we provided the medical students with entire surgical videos (e.g., entire VUA suturing step) to be annotated with the category of the sub-phase (needle handling, driving, and withdrawal) in addition to the start- and end-time of these sub-phases. This resulted in a corpus of video segments and corresponding sub-phases. 

\paragraph{Gesture Annotations} A group of medical students was provided with an instructional manual which delineated the different categories of suturing and dissection gestures. This manual was based on a previously-outlined and peer-reviewed gesture taxonomy\cite{Ma2021}. Once the training phase of the annotation process was complete, we provided the medical students with entire surgical videos (e.g., entire NS step) to be annotated with the category of the gesture in addition to the start- and end-time of the surgical video segment. This resulted in a corpus of video segments (on the order of seconds) and corresponding gestures.

\paragraph{Skills Assessment Annotations} Since we were looking to assess the skills of activities performed during the suturing step, we asked the same set of medical residents who performed the phase annotations (see above) to also annotate such video segments with skill labels. These students were also instructed on how to evaluate suturing skills (low vs. high) based on a scoring system. This scoring system, also known as the \textit{end-to-end assessment of suturing expertise} or EASE, depends on an objective set of criteria\cite{Sanford2022}. Since each video segment was assigned to multiple raters, it had multiple skill assessment labels. In the event of potential disagreements in annotations, we considered the lowest (worst) score. Our motivation for doing so was based on the assumption that if a human rater penalized the quality of the surgeon's activity, then it must have been due to one of the objective criteria outlined in the scoring system, and is thus sub-optimal. We, in turn, wanted to capture and encode this sub-optimal behaviour. 

\subsection*{Reporting summary} 
Further information on research design is available in the Nature Research Reporting Summary linked to this article.

\subsection*{Data availability}
The data from the University of Southern California and St. Antonius Hospital are not publicly available. 

\subsection*{Code availability}
All models were developed using Python and standard deep learning libraries such as PyTorch\cite{Paszke2019}. The code and model parameters will be made publicly available via GitHub.

\bibliography{paper}

\subsection*{Acknowledgments}
We are grateful to Timothy Chu for the annotation of videos with gestures. We also thank Jasper Laca for feedback on presentation. Research reported in this publication was supported by the National Cancer Institute under Award No. R01CA251579-01A1 and a multi-year Intuitive Surgical Clinical Research Grant. 

\subsection*{Author contributions}
D.K. and A.J.H contributed to the conception of the study. R.M and T.H. provided annotations for the video segments. J.N. provided feedback on the manuscript. D.K. contributed to the study design, developed the deep learning models, and wrote the manuscript. C.W. collected data from St. Antonius-Hospital and provided feedback on the manuscript. A.J.H., and A.A. provided supervision and contributed to edits of the manuscript.

\subsection*{Competing Interests}
D.K. is a paid consultant of Flatiron Health. C.W. is a paid consultant of Intuitive Surgical. A.A. is an employee of Nvidia. A.J.H is a consultant of Intuitive Surgical. 

\end{document}


\maketitle


\section*{Supplementary Note 1 - Datasets}

\subsection*{Data Splits}
We use 10-fold Monte Carlo cross-validation in order to evaluate the performance of our deep learning framework. As such, in this section, we outline the training, validation, and test splits for each of those folds across the machine learning tasks described in the main manuscript (surgical phase recognition, gesture classification, and skills assessment). 

\begin{table}[!h]
    \centering
    \begin{tabular}{c | c c c | c c c | c c c}
        \toprule
         \multirow{2}{*}{Fold} & \multicolumn{3}{c}{Training} & \multicolumn{3}{c}{Validation} & \multicolumn{3}{c}{Testing} \\
          & Samples & Videos & Surgeons & Samples & Videos & Surgeons & Samples & Videos & Surgeons \\
         \midrule
0    & 3805          & 63           & 17             & 425         & 7          & 6            & 544          & 8           & 4              \\
1    & 3853          & 63           & 18             & 379         & 7          & 5            & 542          & 8           & 6              \\
2    & 3845          & 63           & 16             & 427         & 7          & 6            & 502          & 8           & 6              \\
3    & 3771          & 63           & 18             & 455         & 7          & 6            & 548          & 8           & 6              \\
4    & 3855          & 63           & 17             & 396         & 7          & 5            & 523          & 8           & 7              \\
5    & 3854          & 63           & 16             & 455         & 7          & 6            & 465          & 8           & 6              \\
6    & 3842          & 63           & 16             & 438         & 7          & 6            & 494          & 8           & 8              \\
7    & 3827          & 63           & 16             & 449         & 7          & 5            & 498          & 8           & 6              \\
8    & 3859          & 63           & 19             & 428         & 7          & 6            & 487          & 8           & 5              \\
9    & 3818          & 63           & 17             & 488         & 7          & 4            & 468          & 8           & 6  \\
         \bottomrule
    \end{tabular}
    \caption{\textbf{Phase recognition - number of samples, videos, and surgeons in the training, validation, and test sets in each fold of the USC VUA dataset.} These data splits are used for the 10-fold Monte Carlo cross-validation.}
    \label{table:vua_phase_usc_data_splits}
\end{table}

\begin{table}[!h]
    \centering
    \begin{tabular}{c | c c c | c c c | c c c}
        \toprule
         \multirow{2}{*}{Fold} & \multicolumn{3}{c}{Training} & \multicolumn{3}{c}{Validation} & \multicolumn{3}{c}{Testing} \\
          & Samples & Videos & Surgeons & Samples & Videos & Surgeons & Samples & Videos & Surgeons \\
         \midrule
0    & 1161          & 66           & 10             & 36          & 11         & 5            & 44           & 12          & 5              \\
1    & 1208          & 65           & 9              & 36          & 10         & 6            & 36           & 10          & 7              \\
2    & 1206          & 65           & 10             & 44          & 11         & 7            & 32           & 12          & 7              \\
3    & 1183          & 67           & 10             & 36          & 11         & 6            & 36           & 11          & 6              \\
4    & 1173          & 65           & 10             & 40          & 11         & 7            & 36           & 10          & 6              \\
5    & 1192          & 64           & 10             & 44          & 11         & 6            & 36           & 11          & 5              \\
6    & 1187          & 66           & 10             & 40          & 10         & 5            & 28           & 11          & 7              \\
7    & 1204          & 64           & 10             & 36          & 11         & 6            & 32           & 10          & 4              \\
8    & 1211          & 66           & 10             & 44          & 11         & 7            & 52           & 12          & 6              \\
9    & 1207          & 66           & 10             & 36          & 10         & 6            & 40           & 11          & 4 \\
         \bottomrule
    \end{tabular}
    \caption{\textbf{Suturing gesture classification - number of samples, videos, and surgeons in the training, validation, and test sets in each fold of the USC VUA dataset.} These data splits are used for the 10-fold Monte Carlo cross-validation.}
    \label{table:suturing_gesture_usc_data_splits}
\end{table}

\begin{table}[!h]
    \centering
    \begin{tabular}{c | c c c | c c c | c c c}
        \toprule
         \multirow{2}{*}{Fold} & \multicolumn{3}{c}{Training} & \multicolumn{3}{c}{Validation} & \multicolumn{3}{c}{Testing} \\
          & Samples & Videos & Surgeons & Samples & Videos & Surgeons & Samples & Videos & Surgeons \\
         \midrule
0    & 1236          & 85           & 15             & 120         & 16         & 10           & 186          & 20          & 9              \\
1    & 1308          & 82           & 15             & 90          & 18         & 10           & 114          & 20          & 10             \\
2    & 1272          & 84           & 15             & 96          & 18         & 10           & 150          & 20          & 11             \\
3    & 1224          & 85           & 15             & 78          & 16         & 12           & 198          & 20          & 9              \\
4    & 1302          & 82           & 15             & 120         & 17         & 12           & 120          & 20          & 11             \\
5    & 1176          & 84           & 15             & 132         & 18         & 7            & 222          & 19          & 9              \\
6    & 1302          & 84           & 15             & 234         & 16         & 8            & 120          & 21          & 8              \\
7    & 1326          & 86           & 15             & 126         & 19         & 11           & 96           & 20          & 8              \\
8    & 1302          & 83           & 15             & 198         & 18         & 8            & 120          & 17          & 8              \\
9    & 1176          & 85           & 15             & 102         & 18         & 10           & 216          & 21          & 10 \\
         \bottomrule
    \end{tabular}
    \caption{\textbf{Dissection gesture classification - number of samples, videos, and surgeons in the training, validation, and test sets in each fold of the USC VUA dataset.} These data splits are used for the 10-fold Monte Carlo cross-validation.}
    \label{table:dissection_gesture_usc_data_splits}
\end{table}

\begin{table}[!h]
    \centering
    \begin{tabular}{c | c c c | c c c | c c c}
        \toprule
         \multirow{2}{*}{Fold} & \multicolumn{3}{c}{Training} & \multicolumn{3}{c}{Validation} & \multicolumn{3}{c}{Testing} \\
          & Samples & Videos & Surgeons & Samples & Videos & Surgeons & Samples & Videos & Surgeons \\
         \midrule
0    & 748           & 63           & 17             & 82          & 7          & 6            & 82           & 8           & 4              \\
1    & 752           & 63           & 18             & 82          & 7          & 5            & 78           & 8           & 6              \\
2    & 778           & 63           & 16             & 44          & 7          & 6            & 90           & 8           & 6              \\
3    & 730           & 63           & 18             & 102         & 7          & 6            & 80           & 8           & 6              \\
4    & 728           & 63           & 17             & 60          & 7          & 5            & 124          & 8           & 7              \\
5    & 774           & 63           & 16             & 46          & 7          & 6            & 92           & 8           & 6              \\
6    & 724           & 63           & 16             & 102         & 7          & 6            & 86           & 8           & 8              \\
7    & 752           & 63           & 16             & 102         & 7          & 5            & 58           & 8           & 6              \\
8    & 754           & 63           & 19             & 86          & 7          & 6            & 72           & 8           & 5              \\
9    & 756           & 63           & 17             & 90          & 7          & 4            & 66           & 8           & 6 \\
         \bottomrule
    \end{tabular}
    \caption{\textbf{Needle handling skills assessment - number of samples, videos, and surgeons in the training, validation, and test sets in each fold of the USC VUA dataset.} These data splits are used for the 10-fold Monte Carlo cross-validation.}
    \label{table:nh_skills_usc_data_splits}
\end{table}

\begin{table}[!h]
    \centering
    \begin{tabular}{c | c c c | c c c | c c c}
        \toprule
         \multirow{2}{*}{Fold} & \multicolumn{3}{c}{Training} & \multicolumn{3}{c}{Validation} & \multicolumn{3}{c}{Testing} \\
          & Samples & Videos & Surgeons & Samples & Videos & Surgeons & Samples & Videos & Surgeons \\
         \midrule
0    & 442           & 63           & 17             & 42          & 7          & 6            & 46           & 8           & 4              \\
1    & 438           & 63           & 18             & 42          & 7          & 5            & 50           & 8           & 6              \\
2    & 432           & 63           & 16             & 44          & 7          & 6            & 54           & 8           & 6              \\
3    & 452           & 63           & 18             & 42          & 7          & 6            & 36           & 8           & 6              \\
4    & 438           & 62           & 17             & 38          & 7          & 5            & 54           & 8           & 7              \\
5    & 448           & 63           & 16             & 30          & 7          & 6            & 52           & 8           & 6              \\
6    & 400           & 63           & 16             & 62          & 7          & 6            & 68           & 8           & 8              \\
7    & 450           & 63           & 16             & 54          & 7          & 5            & 26           & 8           & 6              \\
8    & 408           & 63           & 19             & 48          & 7          & 6            & 74           & 8           & 5              \\
9    & 412           & 63           & 17             & 58          & 7          & 4            & 60           & 8           & 6 \\
         \bottomrule
    \end{tabular}
    \caption{\textbf{Needle driving skills assessment - number of instances (unique videos) in the training, validation, and test sets in each fold of the USC VUA dataset.} These data splits are used for the 10-fold Monte Carlo cross-validation.}
    \label{table:nd_skills_usc_data_splits}
\end{table}